\documentclass{article}
\usepackage{arxiv}

\usepackage[utf8]{inputenc} 
\usepackage[T1]{fontenc}   
\usepackage[colorlinks=true, urlcolor=blue, linkcolor=blue, citecolor=blue,pdfborder={0 0 0}]{hyperref}      
\usepackage{url}           
\usepackage{booktabs}       
\usepackage{amsfonts}     
\usepackage{nicefrac}     
\usepackage{microtype}     
\usepackage{lipsum}	
\usepackage{graphicx}
\usepackage{array}
\usepackage{eqnarray,amsmath}
\usepackage{wrapfig}
\usepackage{mdwlist}
\usepackage{caption}
\usepackage{makecell}
\usepackage{textcomp}
\usepackage{longtable}
\usepackage{multirow}

\title{ State-of-the-Art Augmented NLP Transformer models for direct and single-step retrosynthesis, which overperforms all published approaches}
\date{}

\author{
  Igor V.~Tetko\thanks{Helmholtz Zentrum M{\"u}nchen -- Research Center for Environmental Health (GmbH), Institute of Structural Biology, Ingolst{\"a}dter Landstra{\ss}e 1, D-85764 Neuherberg, Germany} \\
  Institute of Structural Biology\\
  Helmholtz Zentrum M{\"u}nchen,\\
  and BigChem GmbH, \\ 
  Germany, Unterschleißheim \\ 
  \texttt{itetko@bigchem.de} \\ \And
  Pavel Karpov\\
  Institute of Structural Biology\\
  Helmholtz Zentrum M{\"u}nchen,\\
  and BigChem GmbH, \\
  Germany, Unterschleißheim \\
  \texttt{carpovpv@gmail.com} \\ \And 
  Ruud  Van Deursen\\
  Firmenich International SA,\\
  Research\&Development Division, \\
  Switzerland, Geneva \\
  \texttt{ruud.van.deursen@firmenich.com}
  \And
  Guillaume Godin\\
  Firmenich International SA,\\
  Research\&Development Division, \\
  Switzerland, Geneva \\
  \texttt{guillaume.godin@firmenich.com}
}

\begin{document}
\maketitle

\fancyhead{}
\fancyhead[L]{ State-of-the-Art Augmented NLP Transformer models for direct and single-step retrosynthesis \dots}
\fancyhead[R]{A PREPRINT}

\begin{abstract}

We investigated the effect of different training scenarios on predicting the (retro)synthesis of chemical compounds using text-like representation of chemical reactions (SMILES) and Natural Language Processing neural network Transformer architecture. We showed that data augmentation, which is a powerful method used in image processing, eliminated the effect of data memorization by neural networks and improved their performance for prediction of new sequences. This effect was observed when augmentation was used simultaneously for input and the target data simultaneously. The Top-5 accuracy was 84.8\% for the prediction of the largest fragment (thus identifying principal transformation for classical retro-synthesis) for the USPTO-50k test dataset, and was achieved by a combination of SMILES augmentation and a beam search algorithm. The same approach provided significantly better results for the prediction of direct reactions from the single-step USPTO-MIT test set. Our model achieved 90.6\% Top-1 and 96.1\% Top-5 accuracy for its challenging mixed set and 97\% Top-5 accuracy for the USPTO-MIT separated set. It also significantly improved results for USPTO-full set single-step retrosynthesis for both Top-1 and Top-10 accuracies. The appearance frequency of the most abundantly generated SMILES was well correlated with the prediction outcome and can be used as a measure of the quality of reaction prediction.

\end{abstract}

\paragraph{Synposis} Training and application of neural networks with randomized sequences significantly improves direct and retro-synthetic models and can be used to estimate quality of reaction predictions

\section{Introduction}

To synthesize an organic compound is to solve a puzzle with many pieces and potentially several pieces missing. Here, the pieces are single reactions, and finding their sequential combination to create a final product is the retrosynthesis task. 

The success of the logic of organic synthesis developed by E.J. Corey~\cite{Corey} triggered the development of computer programs aiming to find appropriate ways to synthesize a molecule. The first retrosynthesis program LHASA~\cite{LHASA} utilizes a template-based~\cite{SeglerTemplates,coleyTemplates} approach. Every template (rule, synthon) in a curated database of known transformations is sequentially applied to a target molecule, and then sets of reagents are selected according to a specified strategy. Reagents, in turn, undergo the same decompositions until a set of commercially available compounds is found. Retrosynthesis always has multiple routes – a retrosynthetic tree – ending with different starting materials. Thus, a practical algorithm for retrosynthesis has to solve not only the rule acquisition and selection problem but also has capabilities to effectively navigate this tree~\cite{Segler}, taking into account different strategies. These tasks relate directly to artificial intelligence strategies~\cite{BaskinReview,Struble,QsarWithoutBorders}. 

Due to the difficulty of maintaining template databases, most projects dependent on them, including LHASA, did not become widely used tools. The only major exception is, perhaps, the program Synthia\texttrademark (previously CHEMATICA~\cite{CHEMATICA}) which is a successful commercial product. In the Synthia\texttrademark program, rules are automatically extracted from atom-mapped reaction examples~\cite{RouteDesigner}. However, there is an ambiguity in the mapping definition and, more importantly, the automatic rule does not take into account other undefined possible reactive centers in a molecule. Applying such transformations may result in molecules that fail to react as predicted, e.g., 'out-of-scopes' and special care to filter out these cases has to be taken~\cite{Segler}. An alternative approach for the extraction of these rules is to apply a data-driven deep learning technique that corresponds to a machine learning approach where an algorithm (usually in the form of a neural network) is trained on the raw data. After the training finishes, the network contains all the implicitly encoded features (rules) of the corresponding input via its parameters. Works on reaction prediction outcomes~\cite{Schwaller} and retrosynthesis~\cite{Pande,Karpov} showed the feasibility of a symbolic approach, where reactions are written as SMILES~\cite{Weininger} strings as in a machine translation. The product is written in the “source language”, whereas the set of reactants is written in the “target language”. For the “reaction translation” task both languages, however, are SMILES strings, having the same alphabet and grammar. The first works on symbolic (retro)synthesis~\cite{Pande,Kim} were carried out with Seq2Seq~\cite{Sutskever} models following robust and more easy to train Transformer approaches~\cite{Transformer,MolTransformer} that bring state-of-the-art results~\cite{Schwaller,SCORP}. Meanwhile other approaches based on similarity~\cite{Coley}, convolutional~\cite{Ishida,Jin,WLDN}, and graphs show promising results~\cite{RetroLogic,Shi}. 

The SMILES representation of molecules is ambiguous. Though the canonicalization procedure exists~\cite{WeiningerCan}, it has been shown that models benefit from using a batch of random SMILES (augmentation) during training and inference~\cite{Kimber,AugmentationTetko,ESBEN,Swiss}. Recently, such augmentation was also applied to reaction modeling~\cite{Schwaller,MolTransformer,Fortunato,Barzilay}. The augmented (also sometimes called “random”) SMILES are all valid structures with the exception that the starting atom and the direction of the graph enumerations are selected randomly.

In this article, we scrutinize the various augmentation regimes and show that augmentation leads to better performance compared to the standard beam search inference or evaluation of the model under different temperatures.  We clearly mention that our study is to predict single-step and not multi-step retrosynthesis, which has been also targeted using Transformer~\cite{LinTemplateFree,SchwallerHyperGraph}. We show that by using more complicated data augmentation strategies we decrease overfitting~\cite{TetkoStudies} of neural networks and increase their accuracy to achieve top performances for both direct and retro-synthesis. We observe that the harder are the data to train the model, the better it will predict new ones. Moreover, we introduce a new measure MaxFrag accuracy for the prediction of the largest fragment (thus identifying principal transformation for classical retro-synthesis).

\section{Results and discussion}

The baseline dataset contained only canonical SMILES. The other datasets also contained SMILES, augmented as described in the Methods section, p. \pageref{sec:methods}. Four different scenarios were used to augment training set sequences. Namely, we used augmentation of products only (xN), augmentation of products and reactants/reagents (xNF), augmentation of products and reactants/reagents followed by shuffling of the order of reactant/reagents (xNS), and finally mixed forward/reverse reactions, where each retrosynthesis reaction from xNS was followed by the inverse (forward synthesis) reaction (xNM). Only the simplest augmentation xN was used for test sets because no information about reactant/reagents could be used for the retrosynthesis prediction. At least one copy of canonical SMILES for each reaction was present in all augmentation scenarios.

\subsection{Reaction Synthesis Data}
We used a training set filtered from USPTO database~\cite{Lowe} containing 50k reactions classified into 10 reaction types. We used splitting proposed by~\cite{Pande} and divided it into 40k, 5k and 5k reactions for the training, validation, and test sets, respectively. As in the previous study~\cite{Karpov}, after observing that early stopping using the validation set did not improve model test accuracy (the model performance for each of the sets was monotonically increasing with number of iterations, see Fig. \ref{fig:f1s}), we combined the training and the validation sets into a combined training set. The 5k test reactions were predicted only once the model training was finished and were not used at any stage of the model development. In a similar way we joined training and validation sets of USPTO-MIT~\cite{Jin} dataset for direct reaction prediction. In order to provide a more straightforward comparison with results of the previous studies we also reported performances when developing models using only the respective training sets. Moreover a model with the largest published USPTO-full set~\cite{RetroLogic} was also developed.

\subsection{Analysis of canonical datasets}
The development of a model with canonical SMILES (x1) as the training set provided 40.9\% accuracy for prediction of the canonical test set. An attempt to use this model to predict the augmented test set (x5, x10), resulted in much lower Top-1 predictions of 23.3\% and 18.4\%, respectively. This result was to be expected, because the model trained with only canonical sequences was not able to generalize and predict augmented SMILES, which use different styles of molecular representation.

\subsection{Augmentation of products only (xN)}
The augmentation of the products (input data), with just one additional augmented SMILES x2, increased Top-1 accuracy to 43.7\% for the test data composed of canonical sequences. Increasing the number of augmentations in the training set did not increase the Top-1 prediction accuracy. Thus, the augmentation of the training set with just one random SMILES contributed the best performance. This result is in concordance with another study where only one random SMILES was used to augment data~\cite{MolTransformer}.

\subsection{Analysis of character and exact sequence-based prediction accuracy}

\begin{figure}
    \centering
    \includegraphics[width=0.7\textwidth]{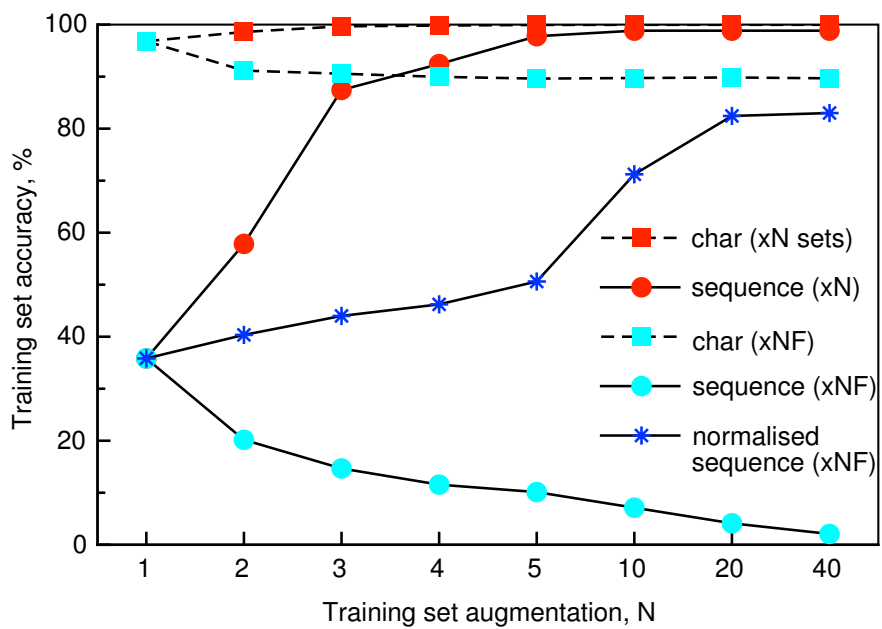}
    \caption{Character and exact sequence based accuracies calculated for the monitoring set. The transformer memorized the target sequences if the target sequences were all canonical SMILES (red dots). It also reasonably predicted the sequence composition for randomized target SMILES (cyan rectangle, dashed) but its performance decreased for prediction of exact full SMILES (cyan circle). The performance normalized by the percentage of canonical sequences increased with the number of augmentations, N, since some of the random sequences were canonical ones.}
    \label{fig:f1}
\end{figure}

To better understand the model training, we also developed several models where approximately 10\% of the dataset did not participate in training but was used to monitor its prediction performance. Different from the test set, which tested the performance of models when predicting a new reaction, the monitoring set tested the ability of the Transformer to predict different SMILES generated for the same reaction. The Transformer was able to recognize different representations of the same reaction. For example, when training x1, the character and exact-sequence-based accuracies when predicting the monitoring sequences were 96.5\% and 34.5\%, respectively. The final performance for the test set, 40.9\%, was higher because some reaction products from the Transformer provided non-canonical SMILES, which were correctly matched after transformation to canonical ones. When using augmented training sequences (x10), the accuracies increased to 99.97\% and 98.9\%, for character and exact sequence-based accuracy, respectively (see Fig. \ref{fig:f1}). The Transformer recognized different representations of SMILES for reactants and reagents of the same training set reaction, and was able to exactly restore the target products which were memorized.  Demonstrably, it was also able to memorize any random sequences. To show this, we used a random SMILES sequence (xNR set in Tables \ref{tbl:s1}, \ref{tbl:s2} and Fig. \ref{fig:f2s}) instead of the canonical sequences as the target for prediction. While this task was more difficult and took more epochs to train, the Transformer was able to perfectly memorize random sequences. Since the SMILES prediction target was random, the Transformer was not able to learn canonicalization rules on how to write the target. Despite this fact, it still calculated a Top-1 prediction accuracy of 26.8\% for the test set which was, however, significantly lower compared to the 42.2\% achieved using the x10 dataset with canonical sequences as the target.

\subsection{Augmentation of reactants and reagents}
A boost of the Transformer performance was observed when, in addition to products i.e. the inputs SMILES, we also augmented the target SMILES, i.e. reactants and reagents. This task was more difficult for the Transformer, which resulted in a drop in both character and sequence based scores for monitoring sequences during the training stage. For example, when using the training dataset with one augmented SMILES, x2F, the character based accuracy dropped to 91.3\%, which was lower than 98.6\% calculated with the x2 dataset composed of canonical product SMILES (Fig. \ref{fig:f1}). For a larger number of augmentations, the character-based accuracy converged to a plateau, e.g. 89.96\% and 89.67\% for the x5F and x20F training sets, respectively. The character-based accuracy was calculated as the percentage of exact matches between target and predicted sequences, e.g. “CCCCN” and “NCCCC” have an accuracy of 80\%, despite being the same SMILES but written from different starting atoms. Thus despite the fact that the Transformer faced a prediction of random SMILES, it was still able to provide a reasonable prediction of their character composition. 

However, of course, the Transformer was not able to predict the exact random product  SMILES. This resulted in a decrease in sequence-based accuracy based on the number of augmentations for xNF training datasets (Fig. \ref{fig:f1}, cyan circle). Still the Transformer was able to predict some of the sequences, which corresponded to the subset of canonical sequences in the monitoring set. Interestingly, the sequence accuracy normalized to the percentage of canonical SMILES in the monitoring sets increased with the number of augmentations since some randomly generated sequences were canonical SMILES.

\subsection{Top-1 performance analysis}
\begin{figure}
    \centering
    \includegraphics[width=0.7\textwidth]{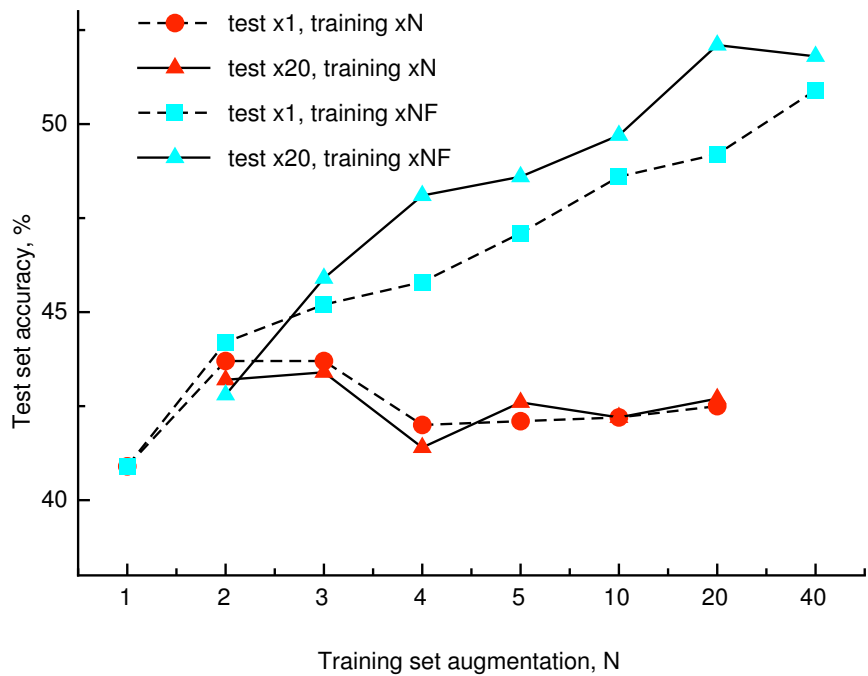}
    \caption{Top-1 performance of models developed with different number of augmentation (shown on x axis) and different augmentation scenarios applied to both test and training sets (red colour: only products were augmented; cyan colour: full reactions were augmented).  The use of the large number of augmentations for the test set (solid lines) improved prediction accuracy for models developed with augmentation of full reactions but did not influence the performance of models where only input data were augmented.}
    \label{fig:f3}
\end{figure}

For augmentations with 1 or 2 random SMILES, the Top-1 prediction performance of the models trained with augmentation of reactants and reagents only, xN, and full reaction augmentation, xNF, were similar. For a larger number of augmentations the models trained with xNF sets had systematically better performances than those developed with xN sets (Fig. \ref{fig:f3}). The training with the x80F set provided the highest Top-1 performance of 52.3\% when this model was applied to the test set generated with x20 number of augmented sequences. While it was possible that further increase in the augmentations could still increase the Top-1 performance, we did not perform such calculations due to limitations on available computational resources.

\subsection{Shuffling order of reactants }
In addition to augmenting the full reaction, we also randomly shuffled the orders of reactants (see xNS set description in Table \ref{tbl:s1} and \ref{tbl:s2}). The effect of this additional data perturbation improved Top-1 performance to 53.1\% for the x20S training dataset applied to the test set with the same number of augmentations (Fig. \ref{fig:f3s}). Further increasing the number of augmentations resulted in the loss of Top-1 prediction accuracy.

\subsection{Shuffling and mixing of retrosynthesis and direct reactions}
The training of retrosynthesis and direct reactions simultaneously could create a mixed representation of latent space and further increase the ability of the Transformer to make  generalizations.  We tested this hypothesis by combining direct and reverse reactions in one training set by reversing the order of product/reactant+reagents and adding a dot to distinguish direct reactions (see e.g., Table \ref{tbl:s2}, x2M augmentation). Contrary to previous analysis, which required 20 augmentations of training set sequences to achieve the highest performance, the mixed dataset achieved it with only 10 augmentations (Fig. \ref{fig:f3s}). Since the mixed dataset also included direct reactions, it had the same number of 19 augmented SMILES per canonical SMILES as in the previous analyses. Thus, this number of augmentations was optimal for the training of the Transformer. A smaller number of augmentations did not allow it to fully explore the data diversity while a larger number created too much noise and made it difficult to learn canonization rules, which were injected by the single canonical sequence. For the x10M training set, the Transformer calculated 52.8\%, which was similar to 53.1\% calculated using the x20S training dataset.

\subsection{Top-5 performance analysis }
This measure provided a relaxed estimation of the performance of the model by measuring if the correct reaction is listed in the Top-5 predicted reactions. Actually, it is questionable whether for retrosynthetic models having the highest Top-1 accuracy is desirable. The goal of a retrosynthetic model is to obtain several precursor suggestions and not exclusively the ones stated in the literature. Moreover, multiple reactions for the same product exist. An example includes the aromatic substitution of an aryl halide (R-X) to an aryl amine (R-NH2) or aryl hydroxide (R-OH). Models with higher Top-n scores do suggest other probable reactions (indeed, all reactions amid Top-n have similar probability) which may correspond to those not reported in the literature for the analysed example. Thus models with higher Top-N scores but with similar Top-1 scores could be interesting for a chemist since they do propose the correct prediction along with similar quality Top-1 reactions.

\begin{figure}
    \centering
    \includegraphics[width=0.7\textwidth]{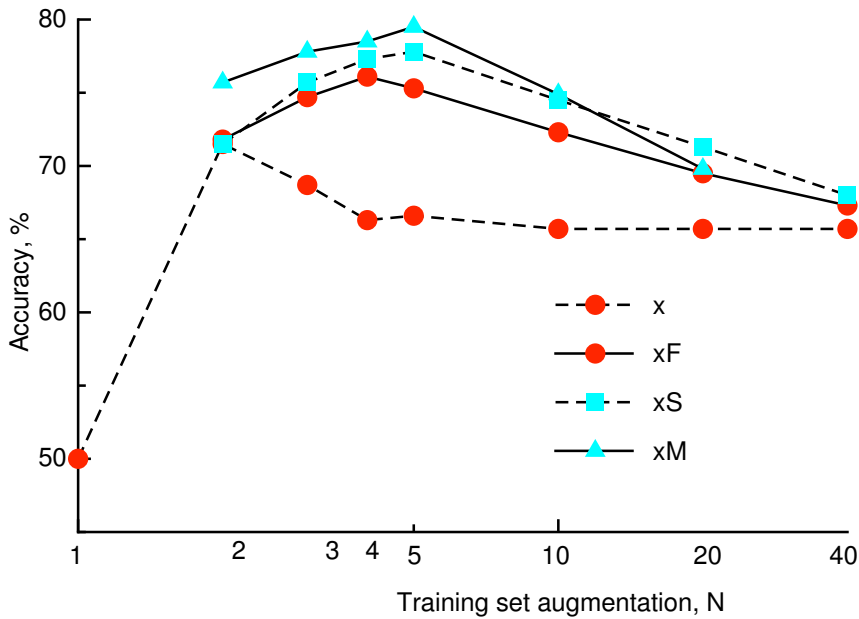}
    \caption{Top-5 performance of transformer models developed with different training set augmentation protocol (See Tables \ref{tbl:s1} \& \ref{tbl:s2}) for prediction of the x20 test set.}
    \label{fig:f5}
\end{figure}

For each number of augmentations, the Top-5 performance generally increased with the number of augmented sequences. The highest Top-5 value was consistently calculated across different scenarios for training sets with 4-5 augmentations only (Fig. \ref{fig:f5}). The highest accuracy, 78.9\%, was calculated for the mixture dataset using the x5M training set augmentation. This number had approximately 1\% higher accuracy than that calculated using the x5S training set (Fig.~\ref{fig:f5}).

\subsection{Reference USPTO-50k model}

For all studies we used a fixed number of epochs N=100. However, we needed to confirm that this was a sufficient number of epochs and to determine if we could calculate better results by training for longer. We selected the model developed with the x5M training set, which provided the highest performance for Top-5 accuracy, and trained it for an additional 400 iterations (in total 500). This additional training improved Top-1 accuracy to 53.3\% while Top-5 performance increased to 79.4\% (Table \ref{tbl:t3}\footnote{The final reference model was built using 500 iterations for the x5M training set. Its reference performance was evaluated using beam size = 5, temperature = 1. The altered parameters are shown for several other application scenarios. For beam = 1 and x1000 augmentations the model calculated 53.7, 80 and 84.3 for Top-1, Top-5 and Top-10 predictions, respectively. This augmentation as well as the one with beam size=10 applied to x100 analysed the same number of predicted sequences. The best results were shown in bold. Larger beam sizes contributed better results for larger Top-n predictions.}) when using beam=5, e.g. the same as in previous analyses.

Further improvement was achieved by using a large number of augmentations, and x100 as the test set. With this setting the model achieved an accuracy of 53.6\% and 80.8\% for Top-1 and Top-5 predictions, respectively. 

\begin{table}[b]
    \centering
    \caption{Analysis of the reference model performance depending on the parameters of the application protocol.}
    \label{tbl:t3}
    \begin{tabular}{p{3cm}p{1.4cm}p{1.4cm}p{1.4cm}p{1.5cm}p{1.4cm}p{1.5cm}p{1.4cm}} \toprule
    \multirow{2}{*}{Apply model setting}  &  \multicolumn{2}{c}{Test set x1} & \multicolumn{2}{c}{Test set x20} & \multicolumn{2}{c}{Test set x100} \\ \cmidrule{2-8}
         & Top-1 & Top-5 & Top-1 & Top-5 & Top-1 & Top-5 & Top-10 \\ \midrule 
    reference accuracy & 48.5 & 72.5 & 53.3 & 79.4 & 53.6 & 80.8 & 85 \\
    temperature, t=1.3 & 49.1 & 67.7 & 52.7 & 77.7 & 53.3 & 78.4 & 83.2 \\ 
    no beam search & 47.7 & 47.7 & 53.3 & 75.3 &  \textbf{53.8} & 78.8 & 81.7 \\
    beam size, beam= 10 & 48.3 & 73.4 & 53.5 & 80 & 53.5 & \textbf{81} & 85.7 \\
    beam size, beam = 44 & 48.3 & 72.5 & 53.5 & 80 & 53.5 & 80.5 & \textbf{85.8} \\ \bottomrule 
    \end{tabular}
\end{table}

\subsection{Influence of temperature}
In our previous study~\cite{Karpov}, we observed that using higher temperatures during beam search increased model accuracy for the Top-1 prediction. It should be mentioned that no augmentation was used in that study. Under the same experimental setup with no augmentation, i.e. when predicting test set composed of only canonical sequences, x1, the Top-1 accuracy of the model increased from 48.3\% to 49.1\% and 49.2\% when using temperatures 1.3 or 1.5, respectively. However, the Top-1 and Top-5 performances for the augmented data (x20) decreased from 53.3\% to 52.7\% and 52.4\%, respectively. For the same test set the Top-5 accuracies also decreased from 79.4\% to 77.7\% and 77.4\% for both temperatures, respectively. Thus, while higher temperatures increased the variability of predictions and thus performance for prediction of canonical sequences, its effect was negative for the augmented data. In particular, it resulted in the lower accuracy of Top-5 predictions.

\subsection{Influence of beam search}
In the above studies we consistently used a beam size of 5 for all analyses. The goal of the beam search was to generate multiple predictions for the same data and thus to better explore the variability of predictions. For example, when using the x20 test set and a beam size of 5, we obtained up to 100 individual predictions, which were used to select the most frequently appearing Top-1 and Top-5 sequences. Increasing the beam size to 10 further improved Top-1 by 0.2 to 53.5\% and Top-5 by 0.6\% to 80\% for the test set. The decrease of the beam size to 3 provided a slightly higher Top-1 score of 53.4\% but decreased the Top-5 to 78.5\% for the same test set. The use of beam size 1 resulted in a Top-1 accuracy of 53.3\% and a reduced Top-5 accuracy of 75.3\% (Table \ref{tbl:t3}). These results were expected: the variation of the beam size slightly influenced the identification of the highest ranked sequence but its smaller number reduced exploration of the space of other Top-reactions for larger $n$.

Both beam search and augmentation increased the number of predicted SMILES which in turn led to better accuracy of model predictions. Thus both of these methods could contribute to the generation of multiple predictions to be used to identify Top-ranked sequences. The maximum size of the beam was restricted by the size of the target vocabulary (number of characters in the target SMILES), which was 44 characters for our dataset. Because of the design of the beam search and because we explicitly excluded duplicated predictions (see section “Analysis of predicted SMILES” as well as Table \ref{tbl:s3}), the dataset used for analysis did not generate duplicated sequences for the same beam search. However, such sequences were indeed generated at different positions of the beam as different representations of the same SMILES. The number of non-unique sequences generated within the same beam search increased with the length of the beam. Interestingly, the use of canonical SMILES as input data contributed to the largest number of unique SMILES, which were 86\%, 82\% and 78\% for beam searches of size 5, 10 and 44, respectively. The use of augmented random SMILES as input contributed smaller numbers of unique sequences, e.g., 42\%, 28\% and 13\% for beam searches of size 5, 10 and 44, respectively. For both types of SMILES some generated SMILES were erroneous and could not be correctly converted by RDKit. Such sequences were excluded from analysis. For large beam sizes, canonical SMILES produced a much bigger percentage of incorrect SMILES, as compared to the use of random SMILES (see Fig. \ref{fig:f4s}). The large difference in the results generated when starting from canonical and random SMILES was also observed for analysis of the percentage of correct predictions for each beam position. In general, the number of erroneous SMILES was low, e.g., on average it was less than 1\% and 3\% for beam search 10, when using augmented and canonical SMILES as input, respectively (Fig. \ref{fig:f4s}). While graph-based methods predict exact chemical structures and thus have 0\% syntactically invalid  SMILES, a few percentage points of incorrectly predicted structures by the Transformer model does not make a large difference to these methods.

The use of canonical SMILES provided (Fig. \ref{fig:f4s}) a higher accuracy for the first beam position, but its  accuracy was much lower for other beams. This was because the Transformer generated canonical SMILES for the canonical input sequences (e.g., 91\% of valid SMILES produced at the position 1 of the beam search for input canonical SMILES were canonical ones) and since only one valid canonical SMILES could be produced, it failed to generate new correct SMILES. Indeed, during the training phase, the Transformer always had a pair of canonical SMILES as input and target sequences. Contrary to that, using augmented SMILES allowed more freedom and allowed it to contribute valid but not necessarily canonical SMILES (e.g., only 33\% of generated SMILES at the position one of the beam search were canonical ones if augmented SMILES were used as input).

The decrease in performance of SMILES generated when using canonical SMILES was one of the main reasons to implement deduplication of data and retain only the first SMILES for the prediction of reactions (see section “Analysis of predicted SMILES”). When deduplication was not performed and all SMILES generated during the beam search were used to rank predictions (compare Tables \ref{tbl:s3} and \ref{tbl:s4}), the Top-1 performances of models were most significantly affected when using only few augmentations, e.g. for the reference model its accuracy dropped from 48.3\% (reference prediction, Table \ref{tbl:t3}) to 47\% but did not change for, e.g. Top-5 performance. In principle, the analysis retaining multiple predicted sequences was based on more data and thus was more stable. Therefore, it could be used when several augmentations and/or large values of Top-n are used for analysis.

As it was mentioned above, both data augmentation and beam search could be used to generate multiple predictions. For the same number of generated sequences, 1000 per SMILES, using a beam = 10 search for the x100 set produced lower accuracy, 53.5\% compared to 53.7\% using augmented data with the x1000 test set without any beam search. The performance of both methods were the same and equal to 53.7\% when the deduplication procedure was not used. However, the beam search contributed to better accuracy, i.e., 81\% vs 80\% and 85.7\% vs 84.3\% compared to the use of augmentation alone for Top-5 and Top-10, respectively. Thus, using beam search allowed a better exploration of data when suggesting several alternative reactions. In any case the augmentation was a very important part of the beam search and for the best performance, both of these approaches should be used simultaneously. We also do not exclude that optimisation of the augmentation may improve its results in the future. Moreover, data augmentation used alone without a beam search contributed superior models to the beam search used without any data augmentation.

\subsection{Accuracy of prediction}
\begin{figure}
    \centering
    \includegraphics[width=0.7\textwidth]{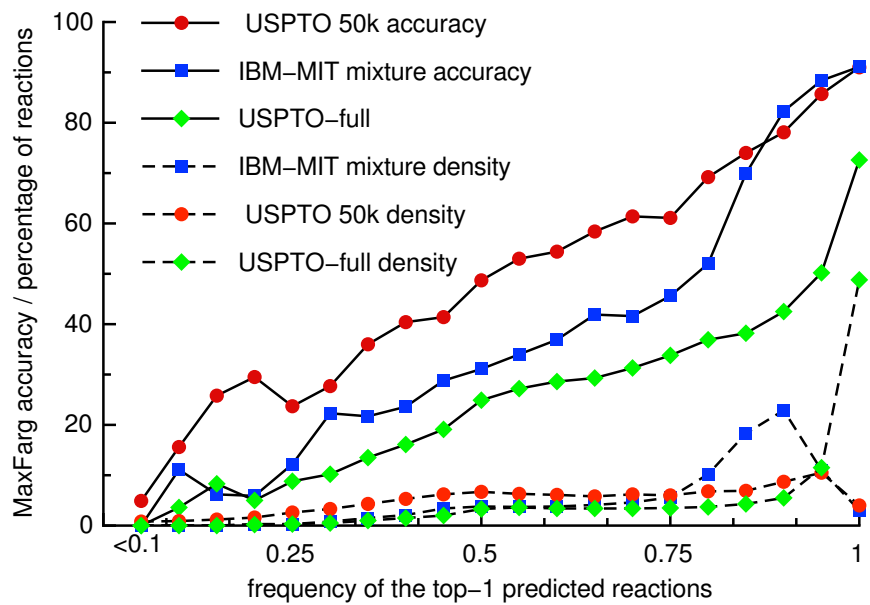}
    \caption{Accuracy and density (fraction of predictions) of the Transformer for MaxFrag Top-1 retrosynthesis accuracy as a function of the frequency of appearance of the Top-1 SMILES in the output of the Transformer for the respective test sets of the models. }
    \label{fig:f4n}
\end{figure}

For some reaction predictions without the use of augmented sequences or position at the beam search the majority of predicted sequences were identical, while for other reactions the Transformer generated as many different SMILES as possible reactants (see Table \ref{tbl:s5}). While the beam generation procedure guaranteed that each prediction had exactly the same sequence of characters, in many cases the Transformer produced multiple non-canonical instances of the same SMILES. The frequency of the appearance of the most frequent (after conversion to the canonical representation) SMILES could, therefore,  indicate the confidence of the Transformer in the prediction. Fig. \ref{fig:f4n} indicated that such frequency (which was calculated on 100x augmented dataset) correlated well with the accuracy of prediction and could be used as a confidence score for the chemist. Indeed, the reactions in which the most frequent SMILES dominated amid all predictions for Top-1 were likely to be predicted correctly. If the most frequent SMILES had low frequencies, such predictions were likely to be incorrect ones. For about 20\% of the most frequent predictions, the accuracy of the retrosynthesis prediction was above 80\% and for 4\% more than 90\%. It should be mentioned, that for a practical implementation which critically depends on the speed, e.g., multistep synthesis, there is no reason to always run all 100 predictions to get the confidence estimations. One can always estimate the probability of the most frequent SMILES and its confidence interval based on a much smaller number of predictions thus decreasing the number of calculations.

As shown in Fig. \ref{fig:f4n}, the same correlations were observed for two other datasets USPTO-MIT and USPTO-Full, which are analysed in the following section. The same approach can be used for Top-n predictions by suggesting one or more plausible pathways for retrosynthesis. An example of such correlations for Top-5 MaxFrag accuracy were shown in Fig.~\ref{fig:f5s}. Moreover, the same approach also predicted the accuracy for the direct synthesis as it was demonstrated at Fig.~\ref{fig:f6s}. It should be mentioned that use of data augmentation is not the only approach to estimate the accuracy of predictions, and other methods based on the  likelihood of the direct reaction prediction were also proposed~\cite{MolTransformer,SchwallerHyperGraph} and were shown to correlate with the accuracy of the predictions. A comparison of such methods is beyond the scope of this study.

\subsection{Analysis of prediction accuracy for different scenarios }

The accuracy of the reference model was about 5\% to 7\% (Top-1) and 10\% (Top-5) higher for reactions without stereochemistry than for those with it (Table \ref{tbl:t2n}\footnote{The classical retro-synthesis accuracy was estimated as the percentage of correctly predicted largest fragments, i.e., “maximum fragment” (MaxFrag) accuracy. The best results are shown in bold.}). 20\% of the reactions in the test set contained molecules with stereochemistry. An increase in the number of augmentations of the test set increased the accuracy of both stereo and non-stereochemical reactions. Stereochemical reactions in the dataset may also suffer from a larger number of annotation errors or/and can have lower prediction scores since such data were underrepresented in the training set. Additionally, for some reactions despite the relative stereochemistry being conserved it may still define confusing information for the model due to the reactant satellite effect. The R/S could be also affected by the way the SMILES was written, e.g. from A to Z or Z to A.

\begin{table}[h]
    \centering
    \caption{Prediction accuracy of the reference model for different subsets of the test set of USPTO-50k using a beam search of size 10.}
    \label{tbl:t2n}
    \begin{tabular}{p{1.7cm}p{1cm}p{1.2cm}p{1.3cm}p{1cm}p{1.2cm}p{1.3cm}p{1cm}p{1.2cm}p{1.3cm}} \toprule
    \multirow{2}{*}{\makecell*[{{p{1.2cm}}}]{Test set augmentation}}  &  \multicolumn{3}{c}{Top-1} & \multicolumn{3}{c}{Top-5} & \multicolumn{3}{c}{Top-10} \\ \cmidrule{2-10}
         & all & stereo (20\%) & no stereo (80\%) & all & stereo (20\%) & no stereo (80\%) & all & stereo (20\%) & no stereo (80\%) \\ \midrule 
    x1 & 48.3 & 44.7 & 49.2 & 73.4 & 67.3 & 74.9 & 77.4 & 71 & 79 \\ 
    x20 & 53.4 & 47.3 & 55 & 80 & 73.3 & 81.9 & 84.2 & 79.2 & 85.4 \\
    x100 & \textbf{53.5} & 47.1 & 55.1 & \textbf{81} & 74.6 & 82.6 & \textbf{85.7} & 81.2 & 86.8 \\
    MaxFrag, x1 & 53.5 & 48.7 & 54.7 & 79.2 & 72.7 &.80.9 & 81.6 & 75.1 & 83.3 \\
    MaxFrag, x20 & 58.5 & 52 & 60.1 & 84.7 & 79 & 86.1 & 88.6 & 83.6 & 89.8 \\
    MaxFrag, x100 & \textbf{58.5} & 51.2 & 60.3 & \textbf{85.4} & 79.4 & 86.9 & \textbf{90} & 85.1 & 91.2  
    \\ \bottomrule 
    \end{tabular}
\end{table}

\subsection{Classical Retro-Synthesis accuracy: recognition accuracy for the largest fragment}

\begin{figure}
    \centering
    \includegraphics[width=\textwidth]{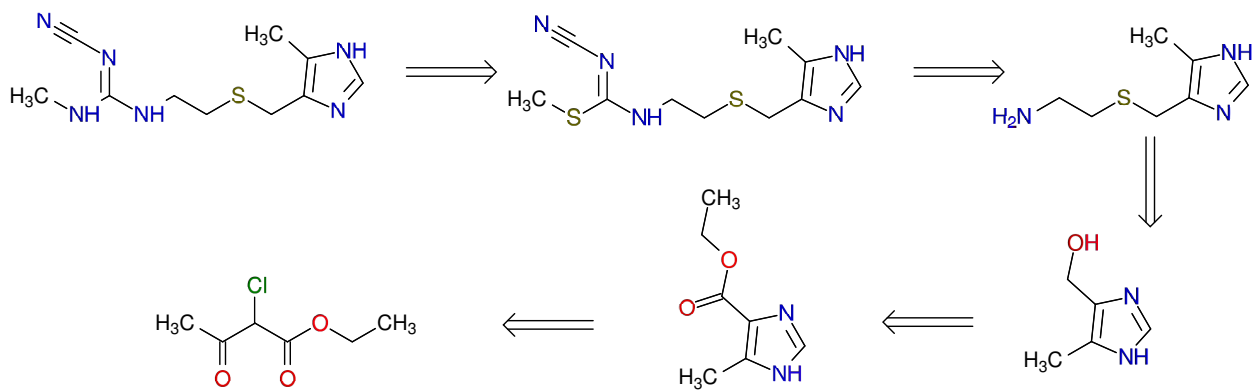}
    \caption{Classical representation of the retrosynthesis of cimetidine focusing on the principal transformations, as is typically written by synthetic chemists (adapted from \href{https://de.wikipedia.org/wiki/Cimetidin}{https://de.wikipedia.org/wiki/Cimetidin} under CC BY-SA 3.0 license). The currently used Top-n accuracy measures also include prediction of other reactants~\cite{Pande,Karpov,SCORP,Coley,RetroLogic,Barzilay}, which may not be necessary for classical methodical retrosynthesis planning.}
    \label{fig:synthesis}
\end{figure}

The prediction of SMILES for retro-synthesis includes exact prediction of the reactants. However, the same reaction performed using different reactants can result in a similar yield. In general the database does not contain all possible reaction conditions to make a given product. Therefore, a prediction of only the main (the largest) reactant can be considered more relevant for retro-synthesis predictions, since we need to first identify the reaction type. Indeed, a chemist generally writes a retrosynthesis by decomposing a target molecule into pieces. This classical procedure, focusing only on main compound transformations, is the minimal information required to get an efficient retrosynthesis route and at the same time all reactions needed (see Fig. \ref{fig:synthesis}). The selection of reaction conditions of the reactions can be considered as a subsequent task.

That is why we decided to consider the recognition of the largest reactant as a new measure of the model performance: Classical Retro-Synthesis Accuracy, i.e.,  the accuracy of prediction of the “Maximal Fragment” (MaxFrag). The MaxFrag was 85.4\% for the Top-5 reaction prediction (Table \ref{tbl:t2n}). The MaxFrag is important to estimate an ability of a system to automatically deduce the correct reaction class. This strategy is orthogonal to explicitly providing reaction class information as input to a model~\cite{RetroLogic}. Adding the reaction class as prior information is equivalent to getting a hint on an exam: this is impractical and also reduces the chance of proposing alternate feasible reactions. Using MaxFrag is more accurate and logical than providing a reaction class as prior information. Besides MaxFrag and Top-n, other scores were proposed to evaluate the success of retro suggestions/reactions, e.g., the matching score by Satoh and Funatsu~\cite{SOPHIA}, the 'in-scope' filter by Segler et al.~\cite{Segler}, and the forward transformer score by Schwaller et al.~\cite{SchwallerHyperGraph}. However, MaxFrag is the easiest and the interpretable one.

\subsection{Retrosynthesis data quality and MaxFrag accuracy}

The use of the classical retro-synthesis accuracy (MaxFrag Top-n) calculated a systematic higher accuracy in comparison to the traditional Top-n scores. To explain this fact, we analysed our datasets and found four types of reactions: non-reagent reactions, one reagent reactions, multiple reagent reactions, and unclear reagent reactions. Non-reagent reactions were reactions that did not work (i.e., A + B -> A). One reagent reactions had only one starting material for the product (A -> P), multiple reagents had multiple starting materials for the same end product (A + B -> P),  and finally unclear reagents where the reaction conditions,  solvent, salts, and so on, were included as reagents (A + B + N -> P, where N were chemicals that did not participate to form the product). Depending on the dataset the proportions of these reaction categories  slightly varied. In the MIT dataset around 0.5\% of reactions were non-reagent reactions, and around 10\% of the reactions were unclear reagent reactions while there were less than 1\% of such reactions in the USPTO-50k dataset. Thus, for the MIT set it would be impossible to fully predict about 10\% of reactions for the retrosynthesis, since they contained chemicals “N” that did not form the reaction, but only conditions, solvent, etc. This more challenging problem of predicting not only the reactants but also the reagents, while still keeping diverse precursor suggestions was addressed elsewhere 34. For the direct synthesis that was not a severe problem since the Transformer could correctly identify and pick-up the interacting components (“A” and “B”) and predict the product. However, the use of Top-n for retrosynthesis is questionable due to the aforementioned problem. The use of MaxFrag accuracy decreased those effects by focusing on the main reagent.  That is why, in our opinion, the MaxFrag score better reflected the chemistry than Top-1 accuracy.

Still there is an unsolved challenge with this score due to the possibilities to synthesize the same products starting from multiple reagents. Both Top-n and MaxFrag Top-n scores were calculated by using the exact match of the predicted and target molecules. But, for example, in the reaction R-R1 + NH3 -> R-NH2 multiples choices of R1, i.e. -OH, -Cl, -Br, -I or -F, would be all correct predictions. The only difference would be the reaction rates and yields, which are not part of the prediction algorithms. Unfortunately the currently used scores, including MaxFrag, cannot yet correctly account for this problem. The problem to some extent could be alleviated by using Top/MaxFrag-n instead of Top/MaxFrag-1 scores: by considering multiple reagents generated by the model, we could also get the one provided in the initial reaction. Thus, the retrosynthesis task is not about getting high Top-1 accuracy. Any classical organic synthesis book, such as the famous Larock’s “Comprehensive Organic Transformations”~\cite{Larock} indicates multiple ways to synthesize chemical compounds and this has to be reflected in the score. The classical retro-synthesis accuracy measured by MaxFrag is a first attempt to better handle those data ambiguities during the validation process and we highly encourage other users to use it. However, in order to enable a comparison with the previous studies we also reported traditional Top-n scores.

\subsection{Analysis of prediction accuracy for different classes}
\begin{figure}
    \centering
    \includegraphics[width=0.7\textwidth]{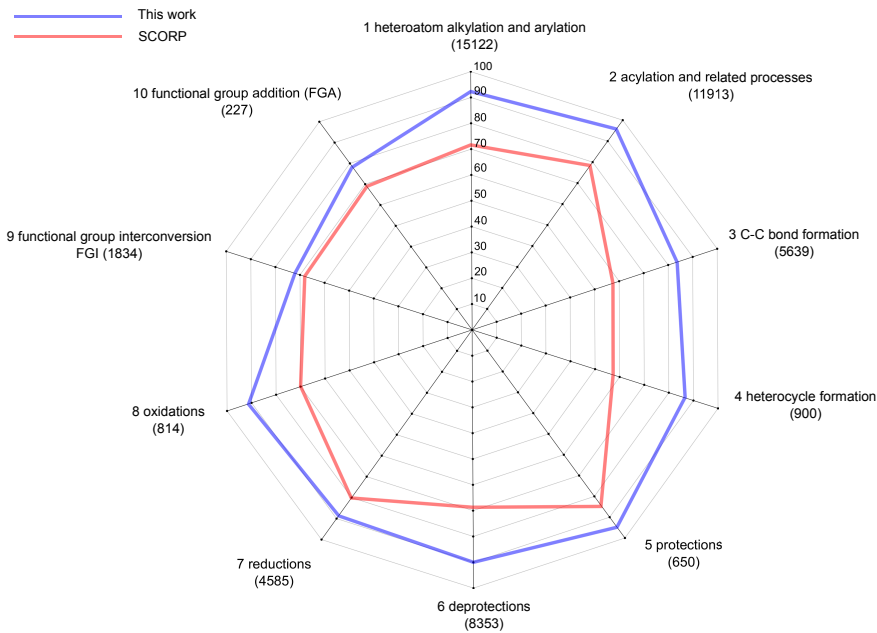}
    \caption{Top-10 accuracy of prediction of different classes of reactions.}
    \label{fig:f5n}
\end{figure}

The original dataset USPTO-50k set~\cite{Pande} provides a reaction type label for every reaction. In total, 10 reaction classes ranging from protection/deprotection, to carbon-carbon bond and heterocycle formation present the most common reactions in organic synthesis. The comparison of accuracy for each class of reactions was presented in Fig. \ref{fig:f5n}. Our best model showed excellent results, outperforming the state-of-the-art Self-Corrected Transformer (SCORP)~\cite{SCORP}. Functional group interconversion and addition, as well as carbon-carbon bond formation were the most difficult for the models to predict. It was not surprising, due to the diverse possibilities for choosing reactions and corresponding reactants for C-C bond creation compared to more straightforward oxidation or protection where the set of groups and reactants is more narrow.

\subsection{Prediction of direct reactions }

The same strategy described in this work was applied to predicting direct reactions from the USPTO-MIT dataset~\cite{Jin}. We used 439k reactions (training and validation set were joined together) as the training set and predicted 40k reactions from the test set by training the Transformer with the same architecture and parameters. The separated and mixed sets were used. In the separated set reactants and reagents were separated with the “>” sign while in mixed set all “>” are substituted with “.” and the order of reactants and reagents was additionally shuffled. The mixed set was more difficult for training since the Transformer had to identify the reaction center from a larger number of molecules. However, such a set better reflected a practical application since separation of data on reactants and reagents in some cases would not be possible without a knowledge of the target product, and thus it did provide a hint to the Transformer about the reaction center. We have removed 316 reactions from the training set where the largest products had length smaller than 5 characters (no reactions were removed from the test set). The Transformer was training using the x5N augmentation protocol for the separated set as well as the x5S and x5M protocols for the mixed set. Since it would be impractical to predict all reagents and reactants for the retrosynthesis task, which was used to additionally augment data in the x5M protocol, only the largest reactant was retained as a target for the reverse reactions.  Augmented test sets were predicted using beam size 10 (Table \ref{tbl:t3n}). For the mixed test set the order of reactants and reagents was shuffled.

\begin{table}[h]
    \centering
    \caption{Prediction accuracy for direct reaction from USPTO-MIT test set using beam size = 10.}
    \label{tbl:t3n}
    \begin{tabular}{p{2.5cm}p{1cm}p{1cm}p{1.2cm}p{1cm}p{1cm}p{1.2cm}p{1cm}p{1cm}p{1.2cm}} \toprule
    \multirow{2}{*}{\makecell*[{{p{1.2cm}}}]{Training set}}  &  \multicolumn{3}{c}{Test set x1} & \multicolumn{3}{c}{Test set x20} & \multicolumn{3}{c}{Test set x100} \\ \cmidrule{2-10}
         & Top-1 & Top-5 & Top-10 & Top-1 & Top-5 & Top-10 & Top-1 & Top-5 & Top-10 \\ \midrule x5N (separated)	& 91.1 &	96.3	&96.7	&91.8	&96.9&	97.3&	91.9&	\textbf{97}&	97.4 \\
    x5S (mixed) &	90&	95.8&	96.2&	90.4&	96.4&	96.9&	90.4&	96.5&	97 \\
    x5M (mixed) &	90&	95.5&	95.7&	90.2&	96.1&	96.5&	90.2&	96.2&	96.8 \\ \bottomrule 
    \end{tabular}
\end{table}

As in previous studies, separation of reagent and reactants with “>” symbols contributed to a model (x5N) with higher prediction scores than for models with mixed sets (x5S and x5M). The additional augmentation of data using retrosynthesis reactions (x5M) did not improve the model. This could be due to the fact that the data for direct reactions were much larger and already contained sufficient information to develop accurate models. While using the x100 test set still contributed better prediction accuracy than using x20, the improvements were in the order of 0.1\% or no improvement at all. Thus the effect of using larger augmentations on model performance reached saturation for the x100 test set.

\subsection{Comparison with published models for direct synthesis using USPTO-MIT set}

The  USPTO-MIT was used as benchmarking for direct synthesis predictions in multiple articles. The AT provided the highest gain in performance for prediction of the more challenging mixed dataset (Table \ref{tbl:t4n}\footnote{The results of the models applied to x100 augmented dataset using beam size = 10. Model was trained on a set of 439k reactions, which combines both the training set of 400k and the validation set of 39k from~\cite{Jin}. The model was trained on the 400k training set to better match performance of previous models.}). Since the model was trained with randomly shuffled augmented data, it was able to generalise very well and provide excellent predictions for the new mixed data. In order to provide a more adequate comparison with previous studies we also developed a model based on exactly the same training data of 400k. Interestingly, the use of a smaller dataset slightly  increased Top-1 performance to 90.6\% but decreased Top-5 performance to 96.1. It should be noted that improvements for direct synthesis look small, i.e. just few percentages. Indeed, the model performance for the direct synthesis increased from 88.6 to 90.6 (Top-1) and 96.1 from 94.2 (Top-5) as compared to the single model reported in ref18. Actually, this is a significant increase in performance since AT decreased the relative errors by 15\% and 30\% for both sets, respectively, if we consider that we can predict direct synthesis with 100\%. In reality we approach the experimental accuracy and further decrease of the errors will be challenging.

\begin{table}[h]
    \centering
    \caption{Comparison of recently published methods for direct synthesis prediction on the USPTO-MIT set.}
    \label{tbl:t4n}
    \begin{tabular}{p{3.4cm}p{1.6cm}p{1.2cm}p{1.6cm}p{1.2cm}p{1.6cm}p{1.2cm}p{1.3cm}} \toprule
\multirow{2}{*}{Model} & \multicolumn{2}{c}{Top-1} & \multicolumn{2}{c}{Top-2} & \multicolumn{2}{c}{Top-5} & \multirow{2}{*}{Reference} \\ \cmidrule{2-7}
& separated & mixed & separated & mixed & separated & mixed & \\ \midrule 
Transformer (single) & 90.4 & 88.6 & 93.7 & 92.4 & 95.3 & 94.2 & \cite{MolTransformer} \\ 
Transformer (ensemble) & 91 & & 94.3 & & 95.8 & & \cite{MolTransformer} \\ 
Seq2Seq & 80.3 & & & & 87.5 & & \cite{Schwaller} \\ 
WLDN & 79.6 & & & & 89.2 & & \cite{Barzilay} \\ 
GTPN & 83.2 & & & & 86.5 & & \cite{GTP} \\
WLDN5 & 85.6 & & & & 93.4 & & \cite{WLDN} \\ 
This work (x100, beam 10) & 91.9 & 90.4 & 95.4 & 94.6 & 97 & 96.5 & \\
AT trained with same training set as in~\cite{Jin} & 92 & 90.6 & 95.4 & 94.4 & 97 & 96.1 \\
\bottomrule 
    \end{tabular}
\end{table}

\subsection{Comparison with published models for retrosynthesis tasks}

\paragraph{USPTO-50k:} The proposed augmentation protocol achieved the best published results on the USPTO-50k dataset (Table \ref{tbl:t5n}). In the previous studies with this set the authors separated data on training, validation and test sets. In all our analyses, since the validation set was not used for model selection and we did not observe the model overfitting~\cite{TetkoStudies} we joined training and validation sets to use all data in order to develop better models.  While we think this is a fair comparison (it is up to the developers of the model to decide on how to best use the available data), we also added results when the model was developed with only the 40k compounds for USPTO-50k set (Table \ref{tbl:t5n}). The accuracies of the models developed with 40k and 45k sets were very similar for the test set. Thus, the data augmentation allowed to compensate for the smaller size of the training set.

\begin{table}[h]
    \centering
    \caption{Comparison of retrosynthesis recently published methods for retrosynthesis prediction on USPTO-50k.}
    \label{tbl:t5n}
    \begin{tabular}{p{2.6cm}p{0.9cm}p{0.9cm}p{0.9cm}p{1cm}p{0.5cm}p{6.7cm}} \toprule
   Model &	Top-1 &	Top-2 &	Top-5 &	Top-10	& Ref	& Comments \\ \midrule 
   Seq2Seq &	37.4 &	&	57.0	&61.7&	 \cite{Pande} &
40/5/5 split; splitting any reactions with multiple products into multiple single product and removal of trivial products \\
Transformer (3*6) &	42.7	&52.5&	69.8&	-	& \cite{Karpov} & 
45/5 split: no validation set was used \\
Transformer (6*8), (self corrected)	& 43.7	&	& 65.2&	68.7&\cite{SCORP} & 
40/5/5 split, reagents from reactants are removed\\
Transformer, augmentation &	44.8 & 57.1	& 57.7	& 79.4 & \cite{Barzilay} & same as in \cite{Pande} \\ 
Similarity-based & 37.3 & 	&63.3 &	74.1 & \cite{Coley} & same as in \cite{Pande} \\
Graph Logic Network & 52.5 & & 75.6	& 83.7	& \cite{RetroLogic} & same as in \cite{Pande,SCORP} \\
G2Gs &	48.9& &	72.5 & 	75.5&\cite{Shi} & same as in \cite{Pande} \\ 
AT	& 53.5	& 69.4	& 81 & 85.7	&  &	same as in \cite{Karpov}. The results of the reference model applied to x100 augmented dataset using beam size = 10.  \\
AT	& 53.2 &	68.1 &	80.5 &	85.2 &		& only 40k samples were used as training set to match the other results. \\ 
AT MaxFrag	& 58.5	& 73	& 85.4 &	90	& & 	same as in \cite{Karpov}. The classical retro-synthesis accuracy was estimated as accuracy for prediction of the largest fragment (MaxFrag). \\ 
AT MaxFrag &	58	&73.4&	84.8& 	89.1& 		&only 40k samples were used as training set to match the other results. \\
\bottomrule 
    \end{tabular}
\end{table}

\paragraph{USPTO-MIT:}  We also analysed the performance of the model at retrosynthesis of the USPTO-MIT set. Compared to USPTO-50k this database also contained multiple reagents and possible catalysts. In our previous analysis (Table \ref{tbl:t3n}), we used the retrosynthesis of the largest fragment as part of the “mix” protocol (x5M), i.e. the products were used as input data contained with “.” to predict the largest reactant (as explained in the supplementary materials, in order to distinguish both direct and retrosynthesis reactions, one of them started with a dot). The dot in front of the SMILES allowed the Transformer to distinguish retrosynthesis from the primary studied direct synthesis reaction. But, of course, the model trained with such data could be also used for retrosynthesis, provided that input data also started with “.”. We also developed a new retrosynthesis model for this set by making it more compatible to USPTO-50k. For this we kept only the 1-2 largest fragments as the targets for retrosynthesis prediction and trained a new model using the x5S protocol. Both models were used to predict the 40k test set which was augmented 100 times. The MaxFrag performances of x5S model, 61.9\% (Top-1), 84.4\% (Top-5) and 86.9\% (Top-10) were very similar to those calculated for the USPTO-50k set (58.5, 85.4 and 90 - see Table \ref{tbl:t5n}). The x5M model, which as aforementioned was a “by-product” of our direct reaction predictions, calculated MaxFrag of 61.1\%, 84.4\% and 88.2\% for the MaxFrag Top1-,Top-5 and Top-10, respectively. Considering that the USPTO-MIT set contained more diverse reactions than USPTO-50k, this result clearly demonstrated the excellent performance of the developed approach and its scalability. The Augmented Transformer (AT) was able to improve its performance for the Top-1 by extracting knowledge from a much larger dataset of reactions.

\paragraph{USPTO-full:} The final testing was done using a USPTO-full set by Dai et al. \cite{RetroLogic}  The authors created a large dataset from the entire set of reactions from USPTO 1976-2016. For reactions with multiple products they duplicated them into multiple ones with one product each. The authors also removed duplications in reactions as well as those with wrong mapping to obtain train/valid/test datasets with 800k/100k/100k sizes. Our analysis identified that some reactions in these sets were still invalid, e.g. contained no products or just single ions as reactants (e.g., US08163899B2,>>[OH2:11], US06048982,CC(=O)OCCCCC[I:22]>>[I-:22], US07425593B2,>>[K:12],  US08114877B2,CC[I:13]>>[I-:13]). We eliminated such reactions as well as those where reactants had less than five atoms in total, since these were unlikely to be correct reactions. This procedure decreased sizes of the train/valid/test sets on average by 4\% to 769k/96k/96k. The AT trained using x5M protocol using the 769k training set calculated the higher performance compared to results from the previous study (Table \ref{tbl:t6n}\footnote{Model was trained using a filtered training set of 769k from \cite{Coley}. Accuracies in parentheses correspond to those recalculated for the test set by assuming that AT failed for all 4\% of excluded reactions. Results for retrosim and neuralsym approaches as reported by Dai et al. \cite{RetroLogic} }). Considering that after the removal of the 4\% erroneous reactions the test dataset was decreased, we also included recalculated performance for it by assuming the worst case scenario: that AT and other tested methods failed for all excluded sequences. Even for this very conservative estimation the AT provided significant improvements compared to previously reported results. The MaxFrag accuracies for USPTO-full were lower compared to that of other analysed sets due to the much higher diversity of this set.

\begin{table}[h]
    \centering
    \caption{Top-k accuracy for retrosynthesis prediction on USPTO-full dataset.}
    \label{tbl:t6n}
    \begin{tabular}{p{4.3cm}p{2.5cm}p{2.5cm}p{2.5cm}p{2.5cm}} \toprule
    &	Retrosim \cite{Coley} & Neuralsym \cite{SeglerTemplates} &	GLN \cite{RetroLogic} &	AT \\ \midrule
    
    Top-1 &	32.8	& 35.8	& 39.3	 & 46.2 (44.4) \\
    Top-2 & & & & 57.2 (54.9) \\ 
    Top-10 & 56.1 & 60.8 & 63.7 & 73.3 (70.4) \\
    MaxFrag Top-1 & & & & 54 \\ 
    MaxFrag Top-2 & & & & 66.3 \\
    MaxFrag Top-5 & & & & 77.3 \\
    MaxFrag Top-10 & & & & 80.6 \\ 
    \bottomrule
   \end{tabular}
\end{table}

Thus for all analysed data sets the AT provided an outstanding performance by consistently and significantly overperforming all previously published models for all statistical performances.

\section{Conclusions and outlook}
This study showed that careful design of the training set was of paramount importance for the performance of the Transformer. Training the model to learn different representations of the same reaction by distorting the initial canonical data eliminated the effect of memorisation and increased the generalisation performance of models. These ideas are intensively used, e.g. for image recognition \cite{Shorten}, and have been already successfully used in the context of several chemical problems \cite{Kimber,AugmentationTetko,ESBEN,Swiss}, including reaction predictions \cite{MolTransformer,Fortunato}, but were limited to the input data. For the first time we showed that application of augmentation to the target data significantly boosts the quality of the reaction prediction. We also showed for the first time that the frequency of predicted SMILES could be used as a confidence metric for (retro)synthesis prediction and can provide quantitative estimation of the most probable reactions amid Top-n predicted outcomes. It is very critical to estimate the quality of the reaction prediction since it could help to better prioritise multi-step retrosynthesis. The developed methodology is unique to the use of augmentation techniques, currently unavailable to GCNs \cite{RetroLogic},  which directly operates with graphs. The estimated accuracy of prediction can help to distinguish reactions, which are difficult to predict, from typo and erroneous reaction data, which will be important to clean up the reaction data and further improve model quality. We also introduced a new MaxFrag measure, classical retro-synthesis accuracy, which in our opinion better reflects the requirements for retrosynthesis analysis.

It should be mentioned that use augmentation was first studied by authors of \cite{MolTransformer}, who introduced Transformer to chemistry and applied it to chemical reactions by using SMILES instead of the text sequences. The augmentation of input data, which was done in that article, provided only a minor improvement of their models. Because of its small impact it was not followed in several other Transformer-based works, including our own study \cite{Karpov,SCORP}.  In this article, we brought an original idea on how to augment chemical data, which provided a significant improvement of the results for all analysed datasets. 

\section*{Supporting Information}
The Supporting Information contains a description of methods, explanation and examples of augmentation protocols, illustration of the procedure of ranking predicted reactions, examples of distributions of predicted SMILES, figures explaining performances of Transformer models. The training and test set (including augmented data), model and model predictions are available at \href{https://github.com/bigchem/synthesis}.

\section*{Acknowledgments}

This study was partially funded by the European Union’s Horizon 2020 research and  innovation program under the Marie Skłodowska-Curie Innovative Training Network European Industrial Doctorate grant agreement No. 676434, “Big Data in Chemistry” and ERA-CVD "CardioOncology" project, BMBF 01KL1710 as well as Intel grant. The article reflects only the author’s view and neither the European Commission nor the Research Executive Agency (REA) are responsible for any use that may be made of the information it contains.  The authors thank NVIDIA Corporation for donating Quadro P6000, Titan Xp, and Titan V graphics cards for this research work. The authors thank Michael Withnall (Apheris AI) and Alli Michelle Keys (Stanford University) for their comments and English corrections as well as Marios Theodoropoulos (University of Geneva) for interesting discussions. We also would like to thank the anonymous reviewers for their insightful and sometimes even provocative comments answering of which significantly increased the value of this study.

\bibliographystyle{naturemag}  
\bibliography{bibliography}

\newpage
\appendix 
\setcounter{table}{0}
\setcounter{figure}{0}
\renewcommand{\thetable}{A\arabic{table}}
\renewcommand{\thefigure}{A\arabic{figure}}
\begin{center}\Large \textbf{Supplementary materials} \end{center}

\section*{Methods}
\label{sec:methods}
\subsection*{Model architecture }

Following our previous study \cite{Karpov} we used the Transformer \cite{Transformer} architecture to train all the models. The key component of the Transformer architecture is a self-attention block equipped with internal memory and attention. During the training phase the block extracts and structures the incoming data, splitting it into memory keys and associated values. Thus, the block resembles a library, where all the books (values) are referred to by an index (keys). On a new request the model calculates the attention to the known keys and then extracts knowledge from the values proportionally. The Transformer shows excellent results not only on (retro) synthesis \cite{Karpov,Schwaller,SCORP} tasks but also on ordinary classification and regression QSAR studies \cite{Swiss}.

The performance of the Transformer was estimated for the prediction of the whole training set after each epoch. The five models with the highest fraction of correctly predicted training set SMILES were stored. As a rule, the stored models correspond to the latest epochs of training. The weights of five stored models were averaged to form the final model, which was used to predict reactions from the test sets.

After several trials, we decided to use a Transformer architecture with 6 layers and 8 heads (6x8), which was used in the original work \cite{Schwaller}. We found that using a smaller architecture with 3 layers and 8 heads (3x8), which was used in our previous study 1, required more epochs to converge and thus longer overall training time to achieve the same performance. We restricted training of the model to 100 epochs to perform model development in a reasonable time and preserve the possibility to compare different augmentation approaches. For the final optimal architectures, we further investigated the effect of training time. 

\subsection*{Influence of the batch size}
The speed of calculations using augmented data was linearly increasing with the dataset size. One epoch using the USPTO-50k set (40k reactions) took 82s on a Titan V. Training of the USPTO-full augmented set (4.3M reactions) took 9514s, i.e. approximately 120 times longer. The use of a larger batch size (in our work we formed batches of length ca. 3000 characters, which approximately corresponded to 12-15 reactions and required about 3.5G of GPU memory for the given Transformer configuration) could increase the speed of calculations. However, we noticed that large batches (i.e., we tried a batch of length 30,000 characters on Tesla V100 with 32GB of memory) could result in a decrease in the speed of convergence. Therefore, for this study we used a batch with 3000 characters.

\subsection*{Beam search}
When generating new SMILES, the Transformer predicted at each step probabilities for all characters from its vocabulary. There are two common approaches to decoding from a linguistic model, such as a Transformer. The first one, a greedy search, always takes the element (symbol, word) with the maximum probability at each step. The second one, beam search 6, tracks in parallel several possible decodings (beam size) and sorts them according to the sums of logarithms of probabilities of each element. Thus, beam search can select those decodings where at one step the element to be chosen has no maximum probability but later symbols have maximum so the overall sum is greater than in greedy search settings. The beam search with n=5 or n=10 beams were used to predict the test set for the majority of analyses performed in this study. As a result of a search using a beam with size n, the Transformer produced up to n SMILES. Because of the generation procedure these were always unique sequences.  Some of them, however, could be errors or could be different representations of the same SMILES. 

\subsection*{Augmentation }

The datasets used in this study comprised both canonical and so-called augmented SMILES.  Both canonical and augmented SMILES were generated using RDKit \cite{RDKIT}. We introduced this SMILES free augmentation method into RDKit at the end of 2018 \cite{Kimber,AugmentationTetko}. The augmented SMILES were all valid structures with an exception that the starting atom and the direction of graph enumerations were selected by chance. The augmentation increased the diversity of the training set. 

The baseline dataset contained only canonical SMILES. The other datasets also contained SMILES augmented as summarized. Four different scenarios were used to augment training set sequences. Sequences were augmented using increasingly complex datasets as shown in Tables \ref{tbl:s1} and \ref{tbl:s2}. Namely, we used augmentation of products only (xN), augmentation of products and reactant/reagents (xNF), augmentation of products and reactants/reagents followed by shuffling of the order of reactant/reagents (xNS), and finally mixed forward/reverse reactions, where each retrosynthesis reaction from xNS was followed by the inverse (forward synthesis) reaction (xNM). One more analysis was performed where the Transformer was asked to predict a fixed random SMILES string.

Only xN were used for augmentations of the test sets because no information about reactant/reagents could be used for the retrosynthesis prediction.

\begin{table}[h]
\caption{Augmentations of analyzed training datasets.}
\label{tbl:s1}
\begin{tabular}{m{1cm}m{14.65cm}} \toprule 
     Dataset & Description \\ \midrule 
     xN & For N=1 the dataset contains canonical SMILES for reactants  and products.  For N>1 in addition to one canonical SMILES, the dataset contains (N-1) instances of the same reaction with augmented SMILES for the products (input data). The SMILES of reactants were canonical. \\ 
xNR & Products are encoded as canonical SMILES, but for reactants only one of possible random SMILES was chosen. \\
xNF & The first instances of each reaction contained canonical SMILES while other (N-1) instances were augmented for both input (products) and output (reactants) data. The order of SMILES in output data was not changed. \\
xNS & Same as xNF but the order of SMILES in reactants was randomly shuffled. \\
xNM & The same as xNS but also contained the same number of inverted (forward synthesis) reactions. The forward reactions started with “.” to distinguish them from retro-synthetic ones.\\ \bottomrule 
\end{tabular}
\end{table}

\begin{longtable}{p{1cm}p{4cm}p{10.2cm}} 
\caption{Examples of data augmentations for two reactions. Canonical SMILES are shown in bold.}
\label{tbl:s2}\\ \toprule 
    \makecell[{{p{1cm}}}]{Dataset} & \makecell[{{p{4cm}}}]{Input (product),\\ output (reactants) data} & \makecell[{{p{10.2cm}}}]{Example} \\ \midrule\endhead\bottomrule\endfoot 
    x0 & \makecell[l]{canonical, \\canonical} & \makecell[l]{\textbf{CC(c1ccc(Br)nc1)N(C)C,CC(=O)c1ccc(Br)nc1.CNC}\\
                                \textbf{O=Cc1cncc(Br)c1,O=C(O)c1cncc(Br)c1}\rule{0mm}{4mm}} \\ \\ 
    x2 & \makecell[l]{canonical,canonical\\random, canonical} & \makecell[l]{\textbf{CC(c1ccc(Br)nc1)N(C)C,CC(=O)c1ccc(Br)nc1.CNC} \\ n1c(Br)ccc(c1)C(N(C)C)C,\textbf{CC(=O)c1ccc(Br)nc1.CNC} \\ \rule{0mm}{4mm}\textbf{O=Cc1cncc(Br)c1,O=C(O)c1cncc(Br)c1} \\ 
c1(cncc(Br)c1)C=O,\textbf{O=C(O)c1cncc(Br)c1}} \\ \\ 

   x2R & \makecell[l]{canonical, fixed random \\ random, fixed random } & \makecell[l]{\textbf{CC(c1ccc(Br)nc1)N(C)C, c1cc(Br)ncc1C(=O)C.CNC} \\ 
n1c(Br)ccc(c1)C(N(C)C)C, c1cc(Br)ncc1C(=O)C.CNC \\ 
\rule{0mm}{4mm}\textbf{O=Cc1cncc(Br)c1}, c1c(cncc1C(=O)O)Br \\
c1(cncc(Br)c1)C=O, c1c(cncc1C(=O)O)Br} \\ \\ 

x2F & \makecell[l]{canonical, canonical \\ random, random} & \makecell[l]{\textbf{CC(c1ccc(Br)nc1)N(C)C, CC(=O)c1ccc(Br)nc1.CNC} \\
n1c(Br)ccc(c1)C(N(C)C)C, CC(=O)c1ccc(nc1)Br.CNC \\ 
\rule{0mm}{4mm}\textbf{O=Cc1cncc(Br)c1, O=C(O)c1cncc(Br)c1} \\ 
c1(cncc(Br)c1)C=O, c1c(cncc1C(=O)O)Br} \\

x3S & \makecell[l]{canonical, canonical \\ 
random, shuffled \\ 
random, shuffled \\ 
} & \makecell[l]{\textbf{CC(c1ccc(Br)nc1)N(C)C,CC(=O)c1ccc(Br)nc1.CNC}\\ 
n1c(Br)ccc(c1)C(N(C)C)C,CNC.CC(=O)c1ccc(nc1)Br \\ 
CN(C(c1ccc(Br)nc1)C)C,CNC.c1cc(Br)ncc1C(O)C \\ 
\rule{0mm}{4mm}\textbf{O=Cc1cncc(Br)c1,O=C(O)c1cncc(Br)c1} \\
c1(cncc(Br)c1)C=O,c1c(cncc1C(=O)O)Br \\
n1cc(cc(c1)C=O)Br,OC(=O)c1cncc(c1)Br} \\ \\ 

x2M & \makecell[l]{canonical, canonical\\
.canonical, canonical\\
random, shuffled\\
.shuffled. random\\
} & \makecell[l]{\textbf{CC(c1ccc(Br)nc1)N(C)C,CC(=O)c1ccc(Br)nc1.CNC} \\
.CC(=O)c1ccc(Br)nc1.CNC,CC(c1ccc(Br)nc1)N(C)C \\
n1c(Br)ccc(c1)C(N(C)C)C,CNC.CC(=O)c1ccc(nc1)Br \\ 
.CNC.CC(=O)c1ccc(nc1)Br,n1c(Br)ccc(c1)C(N(C)C)C \\
\rule{0mm}{4mm}\textbf{O=Cc1cncc(Br)c1,O=C(O)c1cncc(Br)c1}\\
.O=C(O)c1cncc(Br)c1,O=Cc1cncc(Br)c1\\
c1(cncc(Br)c1)C=O,c1c(cncc1C(=O)O)Br\\
.c1c(cncc1C(=O)O)Br,c1(cncc(Br)c1)C=O} \\
\end{longtable}

\section*{Analysis of predicted SMILES}
\label{sec:analysis-smiles}
The beam search was used to infer n=5 (or more) reactant sets from the model for each entry in the test file. The SMILES predicted within the same beam search were sorted in the decreasing order of their probabilities. Predictions containing erroneous SMILES representations, which could not be processed by RDKit, were discarded. The remaining predictions were converted to canonical SMILES. In cases where the predicted reaction contained several disconnected SMILES, they were sorted to have the same representation. If two or more identical predictions were found for the same input only the first prediction was kept: in this way we deduplicated reactions predicted for the same input data. For augmented test datasets, SMILES predicted for the same reaction were accumulated and those with the largest number of occurrences were selected as the Top-ranked. If exactly the same number of predictions were found for two or more SMILES, the weights of the SMILES were set to be inversely proportional to their relative position in the respective beam search. Precisely, to rank predictions we used the following formula

\begin{equation}\label{eq:rank}
    rank(SMILES) = \sum_{n\in[{0,augmentations})} \sum_{i\in[1,beam]}\frac{\delta(SMILES_{n,i}, TARGET)}{1.0 + 0.001 * i} 
\end{equation}

where the first sum was over canonical (n=0) and augmented SMILES for the same input reaction. When the target canonicalized SMILES was equal to the predicted canonicalized SMILES at position i of the beam search for augmentation n, $\delta$=1. Otherwise, if predicted and target SMILES did not coincide, $\delta$=0. The term 0.001*i was used to weight the predicted SMILES to be inversely proportional to its position in the beam search (see also Table \ref{tbl:s3} and \ref{tbl:s4}).

\subsection*{Top-n performance accuracy}

For the analysed input reaction we received a set of generated canonical SMILES (contributed both by beam search and augmentation procedure), which were ranked as explained above. If any of these Top-n sequences coincided with the target canonical SMILES for the analysed reaction, the prediction was considered to be the correct one. The Top-n accuracy was the ratio of the number of correct predictions to the total number of sequences in the test set.

\begin{table}[h]
\caption{Illustration of a procedure used to rank predicted reactions.}
\label{tbl:s3}
\begin{tabular}{p{3cm}p{3cm}p{9.2cm}} \toprule 
Step & Input SMILES & Beam1,Beam2,Beam3  \\ \midrule 
Initial prediction & \makecell[l]{SMILES\_CAN \\ SMILES\_AUG1 \\ SMILES\_AUG2} & 
\makecell[l]{CC(C),C(C)CC,C(N)N\\ CNN,CCC,CC=\\CC.CCC,CCC.CC,C\# } \\ \\ 
Canonicalisation, sorting and error detection & \makecell[l]{SMILES\_CAN \\ SMILES\_AUG1 \\ SMILES\_AUG2} & 
\makecell[l]{CCC,CCCC,CNN \\CNN,CCC,error\\CC.CCC,CC.CCC,error} \\ \\ 
Elimination of duplicates and erros & \makecell[l]{SMILES\_CAN \\ SMILES\_AUG1 \\ SMILES\_AUG2} & 
\makecell[l]{CCC,CCCC,CNN\\CNN,CCC\\CC.CCC} \\ \\
Enumeration & \makecell[l]{SMILES\_CAN \\ SMILES\_AUG1 \\ SMILES\_AUG2} &  
\makecell[l]{CCC(0),CCCC(1),CNN(2) \\ CNN(0),CCC(1) \\ CC.CCC(0)} \\ \midrule 
Ranks, see Eq. \ref{eq:rank}. & & 
\makecell[l]{
CCC = [1] +[ 1/(1+1./1000)] + [0]  = \textbf{1.999} \\
CNN = [1/(1+2./1000)] + [1] + [0] = 1.998\\ 
CC.CCC =[ 0] + [0] + [1] = 1\\ 
CCCC = [1/(1+1./1000)] + [0] + [0] = 0.999\\ The Top-2 ranked predictions are CCC and CNN.} \\ 

 \bottomrule 
\end{tabular}
\end{table}

\begin{table}[h!]
\caption{Illustration of procedure used to rank predicted reactions when using multiple predictions within the same beam.}
\label{tbl:s4}
\begin{tabular}{p{3cm}p{3cm}p{9.2cm}} \toprule 
Step & Input SMILES & Beam1,Beam2,Beam3  \\ \midrule 
Initial prediction & \makecell[l]{SMILES\_CAN \\ SMILES\_AUG1 \\ SMILES\_AUG2} & 
\makecell[l]{CC(C),C(C)CC,C(N)N\\ CNN,CCC,CC=\\CC.CCC,CCC.CC,C\# } \\ \\ 
Canonicalisation, sorting and error detection & \makecell[l]{SMILES\_CAN \\ SMILES\_AUG1 \\ SMILES\_AUG2} & 
\makecell[l]{CCC,CCCC,CNN \\CNN,CCC,error\\CC.CCC,CC.CCC,error} \\ \\ 
Elimination of duplicates and erros & \makecell[l]{SMILES\_CAN \\ SMILES\_AUG1 \\ SMILES\_AUG2} & 
\makecell[l]{CCC,CCCC,CNN\\CNN,CCC\\CC.CCC} \\ \\
Enumeration & \makecell[l]{SMILES\_CAN \\ SMILES\_AUG1 \\ SMILES\_AUG2} &  
\makecell[l]{CCC(0),CCCC(1),CNN(2) \\ CNN(0),CCC(1) \\ CC.CCC(0)} \\ \midrule 
Ranks, see Eq. \ref{eq:rank}. & & 
\makecell[l]{
CCC = [1] +[ 1/(1+1./1000)] + [0]  = \textbf{1.999} \\
CNN = [1/(1+2./1000)] + [1] + [0] = 1.998\\ 
CC.CCC =[0] + [0] + [1] + [1/(1+1./1000)] = \textbf{1.999}\\ 
CCCC = [1/(1+1./1000)] + [0] + [0] = 0.999\\ The Top-2 ranked predictions are CCC and  CC.CCC.} \\ 

 \bottomrule 
\end{tabular}
\end{table}

The SMILES strings with the largest weights and thus those that appeared most frequently amidst the first sequences within the beam predictions were selected as the Top-ranked. The Top-1 and Top-5 SMILES were used to estimate the prediction performances of models.

\subsection*{Analysis of stereochemistry-free datasets}
About 20\% of the reactions in the training and test sets had molecules with stereochemistry. The stereochemistry was encoded in SMILES with “/”,”\textbackslash”,”@” and “@@” characters. However, a number of practical projects have relaxed stereochemistry requirements. Therefore, we separately reported the performance of the models for datasets with and without stereo-chemical information.

\subsection*{Character and exact sequence performance during training }

During the model training, we calculated character-based performance, which corresponded to the number of exactly predicted characters for the target SMILES, as well as exact sequence accuracy indicating the number of correctly predicted exact sequences. Both of these measures were approximations of the final accuracy, for which the predicted SMILES were first converted to canonical ones and only after were compared to the target values.

\begin{table}[h!]
\caption{Examples of distributions of predicted SMILES.}
\label{tbl:s5}
\begin{tabular}{p{7cm}p{5cm}p{3cm}} \toprule 
Reaction & Frequency of SMILES & Ratio of the most frequent to all SMILES  \\ \midrule 

CCOC(=O)C1CCCN(C(=O)COc2ccc(-c3ccc(C\#N)cc3)cc2)C1>>CCOC(=O)C1CCCNC1. N\#Cc1ccc(-c2ccc(OCC(=O)O)cc2)cc1 & 
926* 51 7 6 2 1 1 1 1 1 1 1 & 926/999 = 0.93 \\ \\

CCCCC(=O)O>>CCCCC(=O)OC(=O)CCCC & 
203 112 107 98 57 19 16 13 12 12 12 12 11 11 11 11 11 11 11 11 11 11 8 8 8 8 6 6 6 6 6 6 6 6 5 5 5 5 5 5 5 5 5 4 4 4 4 4 3 3 3 3 3 3 3 3 3 3 3 3 3 3 3 3 3 3 2 2 2 2 2 2 1 1 1 1 1 1 & 203/999 =  0.2 \\ \\ 
\multicolumn{3}{m{16cm}}{Star indicates the correctly predicted reaction. For the first reaction the most frequent SMILES was predicted 926 times or 93\% of all predictions. For this SMILES the Transformer was very confident in the outcome of retrosynthesis, which it correctly predicted. For the second reaction the Transformer generated 78 different SMILES and the Top-1 SMILES appeared only in 20\% of all predictions. For this reaction the Transformer failed to predict the correct SMILES at all.}
\\ \bottomrule 
\end{tabular}
\end{table}

Unless otherwise noticed, the results presented in the supplementary Figures were calculated using the USPTO-50k set.

\begin{figure}
    \centering
    \includegraphics[width=0.9\textwidth]{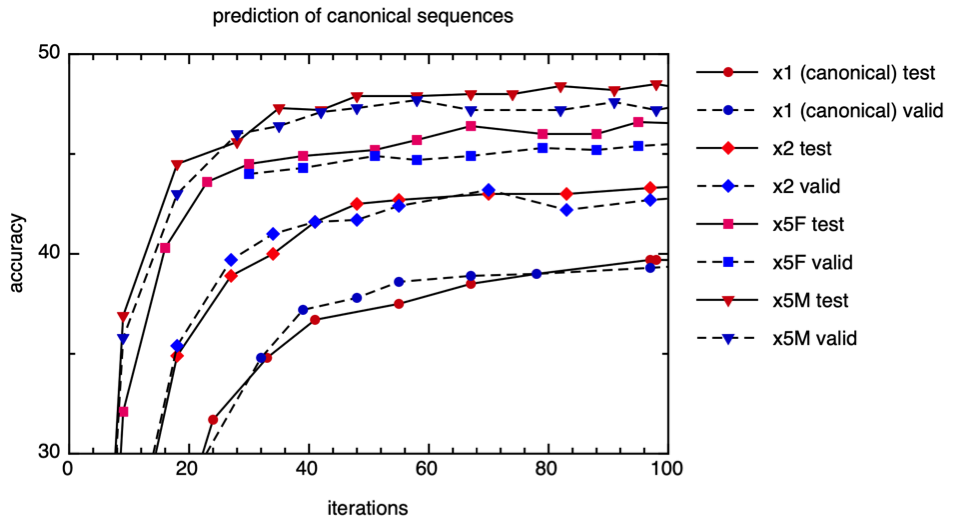}
    \caption{Top-1 accuracies calculated for models developed with different augmentation scenarios and using 40k sequences as the training set. All models were applied to the x1 (canonical) test and validation set as they were defined in 10. As we can see the performance of models is similar for both training and validation sets and it is monotonically increasing with the number of iterations. This observation was the main motivation to join training and validation sets as a single set, which was used for model development.}
    \label{fig:f1s}
\end{figure}

\begin{figure}
    \centering
    \includegraphics[width=0.7\textwidth]{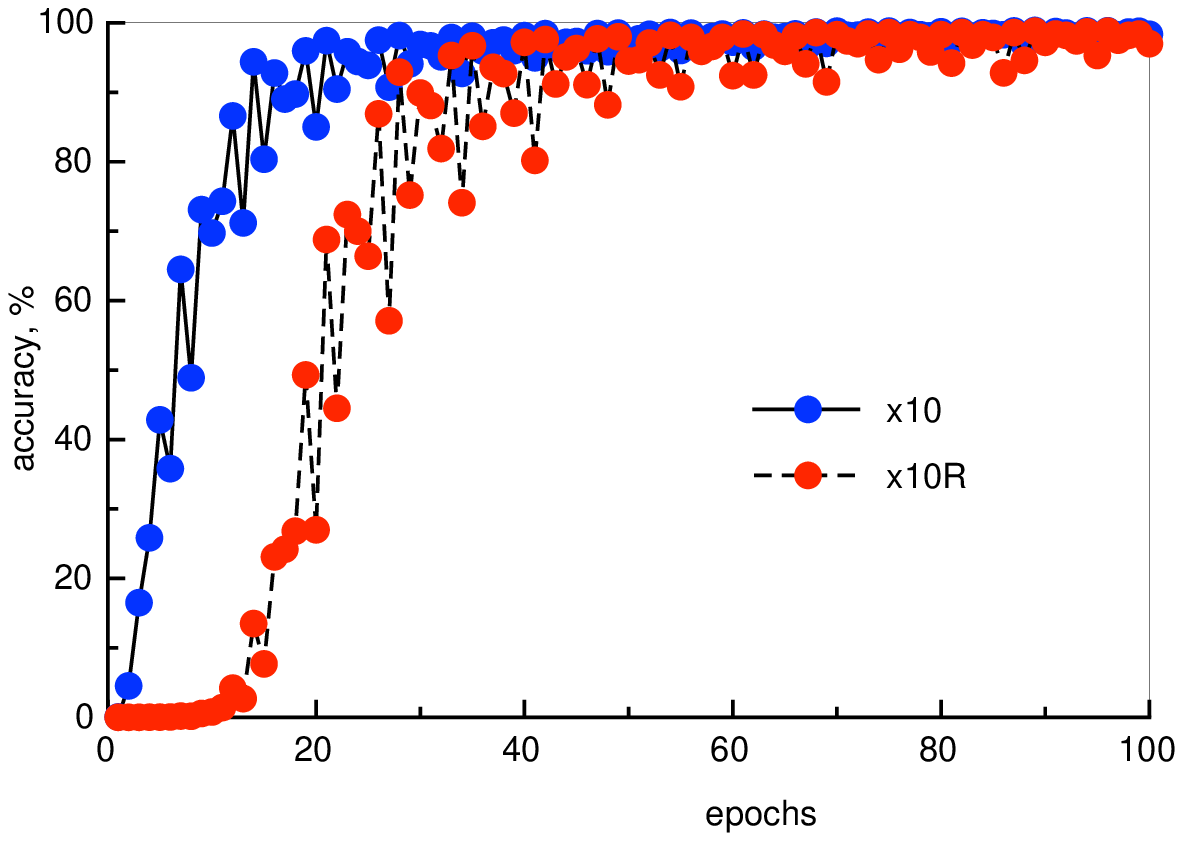}
    \caption{Monitoring set accuracy (measured as a character accuracy) of Transformer for prediction of canonical (x10) and random (x10R) SMILES for USPTO-50k set (see Table 1 and 2 for explanation of used abbreviations).}
    \label{fig:f2s}
\end{figure}

\begin{figure}
    \centering
    \includegraphics[width=0.7\textwidth]{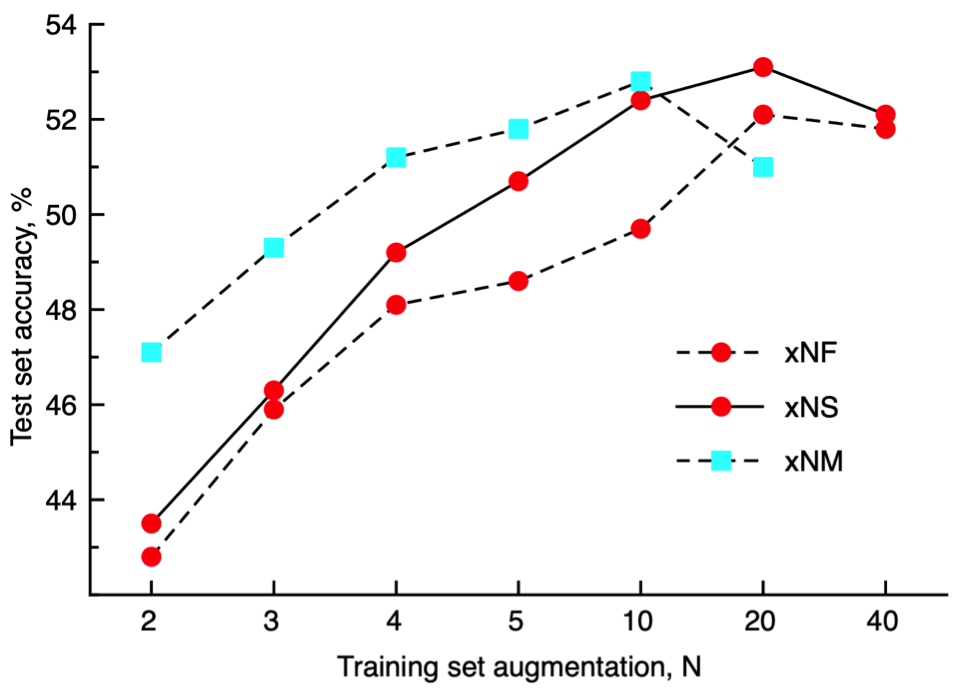}
    \caption{Top-1  full-sequence retrosynthesis accuracies calculated for models developed with different augmentation scenarios for USPTO-50k training set. All models were applied to the x20 test set.}
    \label{fig:f3s}
\end{figure}

\begin{figure}
    \centering
    \includegraphics[width=0.7\textwidth]{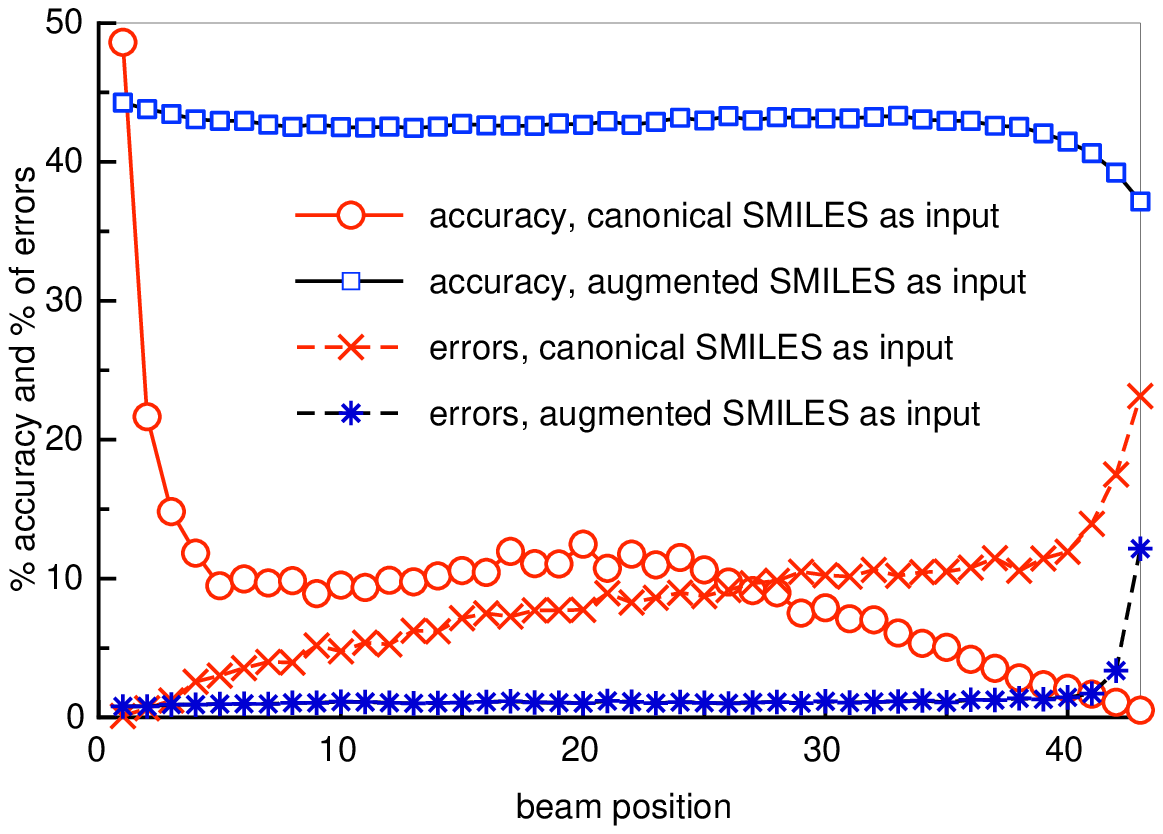}
    \caption{Accuracy of prediction of SMILES generated at the respective position of the beam search using the largest beam size=44. The results were calculated for test set prediction using the model trained with 500 iterations on the USPTO-50k set. The use of canonical SMILES as input produced the highest accuracy (48.3\%) for the first beam, which degraded for other positions of the beam while use of augmented SMILES provides about 44\% correct predictions, which is slowly decreasing with the increase of the beam position. The number of erroneous SMILES is increasing with the beam position for both types of SMILES, but it was significantly higher for predictions when using canonical SMILES as input.}
    \label{fig:f4s}
\end{figure}

\begin{figure}
    \centering
    \includegraphics[width=0.7\textwidth]{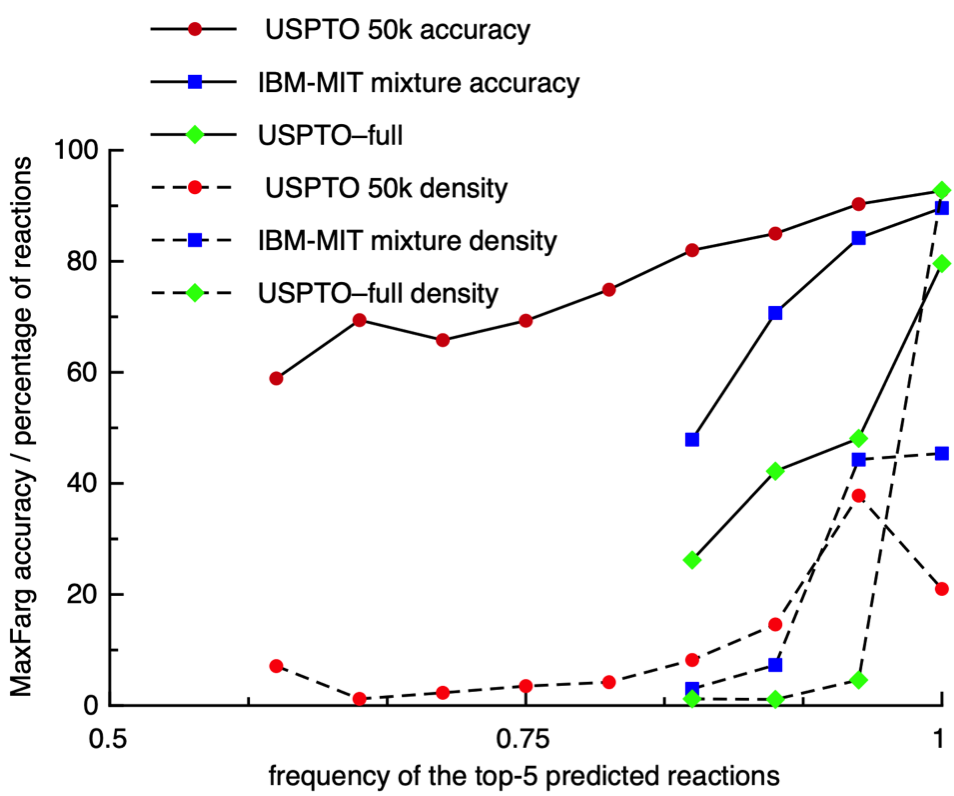}
    \caption{Accuracy and density (fraction of predictions) of the Transformer for MaxFrag Top-5 retrosynthesis accuracy as a function of the frequency of appearance of five most frequent SMILESes in the output of the Transformer (see also Fig. 4 in the article). Due to a small number of samples and high variability of data the average accuracy is shown for each first left datapoint for the same or smaller frequencies. For example for USPTO 50k set the accuracy of 58.9\% for frequency 0.6 was calculated by averaging the MaxFrag accuracies for SMILES with frequencies $\leq$ 0.6. There were 7.1\% of such predictions in the test dataset.}
    \label{fig:f5s}
\end{figure}

\begin{figure}
    \centering
    \includegraphics[width=0.7\textwidth]{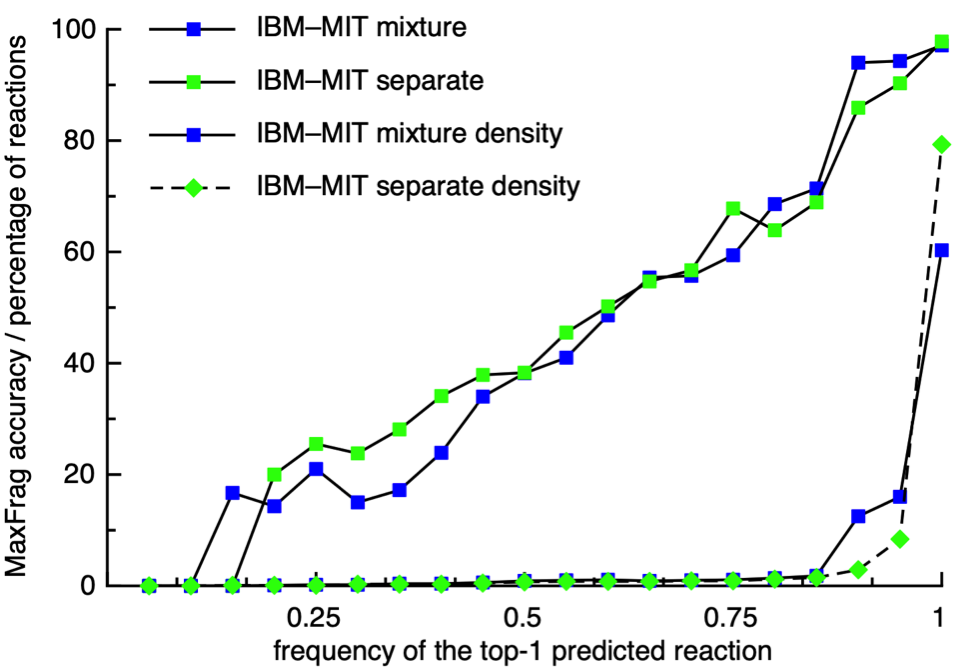}
    \caption{Accuracy and density (fraction of predictions) of the Transformer for MaxFrag Top-1 direct synthesis accuracy as a function of the frequency of appearance of the Top-1 SMILES in the output of the Transformer for the respective test sets of the models (see also Fig. 4 in the article). }
    \label{fig:f6s}
\end{figure}

\subsection*{Calculation of the decrease of the relative errors}

Let us assume that we can (theoretically) get 100\% accuracy.  Top-5 error of the previous model for the mixed set was 100-94.2 = 5.8\%. The error of our model is 100 - 96.1 = 3.9\%. The relative decrease of the error is (5.8-3.9)/5.8 = 32.7\%.  

~
\vfill

\end{document}